\documentclass[a4wide]{article}

\usepackage[margin=1.5in]{geometry}
\usepackage{amssymb, amsmath, amsopn}
\usepackage{times,mathptmx}
 \usepackage{mathptmx}
 \usepackage{helvet}
 \usepackage{courier}
 \usepackage{makeidx}
 \usepackage{multicol}
 \usepackage{footmisc}
 \usepackage[justification=centering]{caption}
 \usepackage{amssymb, amsmath}
 \usepackage{graphicx}
 \usepackage{caption}
 \usepackage{subfig}
 \usepackage{xspace}
 \usepackage{color}
 \usepackage{array}
 \newcolumntype{P}[1]{>{\centering\arraybackslash}p{#1}}
 \usepackage{verbatim}
 \usepackage{imakeidx, epsfig,url}
\usepackage{algorithm}
\usepackage[algo2e,ruled,vlined]{algorithm2e}
\usepackage{xspace}
\usepackage[numbers,sort&compress,longnamesfirst,sectionbib]{natbib}
\usepackage{balance}

\usepackage{graphicx}
\usepackage{rotating}
\usepackage{amssymb}
\usepackage{graphicx}
\usepackage{url} 

\usepackage{color}
\usepackage{xspace}
\usepackage{blindtext}
\usepackage{multirow, rotating}
\usepackage{algorithm,algpseudocode,booktabs,amsmath}
\makeatletter
\renewcommand{\ALG@beginalgorithmic}{\small}
\makeatother

\usepackage{hyperref}
\usepackage{balance}
\usepackage{tikz}
\usetikzlibrary{shapes,decorations,patterns,positioning}
\usetikzlibrary{plotmarks}
\graphicspath{{/plots/}{plots/}}
\newcommand{\ignore}[1]{}

\newcommand{\tmax}{t_{\max}}
\newcommand{\eaasym}{(1+1)~EA$_{\text{asym}}$\xspace}
\newcommand{\rwalk}{EA-UniformWalk\xspace}
\newcommand{\bwalk}{EA-BiasedWalk\xspace}
\newcommand{\arwalk}{EA-AsymUniformWalk\xspace}
\newcommand{\abwalk}{EA-AsymBiasedWalk\xspace}

\usepackage{url}

\begin{document}
\title{Evolutionary Image Transition Based on Theoretical Insights of Random Processes}

\author{
Aneta Neumann\\
Optimisation and Logistics\\
School of Computer Science\\
The University of Adelaide\\
Adelaide, Australia
\and 
Bradley Alexander\\
Optimisation and Logistics\\
School of Computer Science\\
The University of Adelaide\\
Adelaide, Australia
\and 
 Frank Neumann\\
 Optimisation and Logistics\\
 School of Computer Science\\
 The University of Adelaide\\
 Adelaide, Australia
}

\maketitle

\begin{abstract}
Evolutionary algorithms have been widely studied from a theoretical perspective. In particular, the area of runtime analysis has contributed significantly to a theoretical understanding and provided insights into the working behaviour of these algorithms. We study how these insights into evolutionary processes can be used for evolutionary art. We introduce the notion of evolutionary image transition which transfers a given starting image into a target image through an evolutionary process. Combining standard mutation effects known from the optimization of the classical benchmark function OneMax and different variants of random walks, we present ways of performing evolutionary image transition with different artistic effects.
\end{abstract}

\sloppy

\section{Introduction}\label{sec:intro}

Evolutionary algorithms (EAs) have been successfully used in the areas of music and art~\cite{DBLP:conf/ncs/2008evolution}. In this application area the primary aim is to evolve artistic and creative outputs through an evolutionary process~\cite{DBLP:conf/ncs/2008evolution,citeulike:12541313,DBLP:conf/evoW/VinhasACEM16}. 
The use of evolutionary algorithms for the generation of art has attracted strong research interest. 
Different representations have been used to create works of greater complexity in 2D and 3D~\cite{Todd:1994:EAC:561831}, and in image animation \cite{DBLP:conf/siggraph/Sims91,DBLP:conf/evoW/Hart07,DBLP:conf/ncs/Draves08}.  
The great majority of this work relates to using evolution to produce a final artistic product in the form of a picture, sculpture or animation. 

The focus of study in this paper is how EA processes can be mirrored in image transitions. Past work studying the use of EAs for  image transitions includes work by Sims~\cite{DBLP:conf/siggraph/Sims91} which described methods for cross-dissolving of images by changes in an expression genotype. Furthermore, deep neural networks have recently been used to create artistic images~\cite{DBLP:journals/corr/GatysEB15a}.
Banzhaf~\cite{DBLP:conf/eps/GrafB95} used interactive evolution to help determine parameters for image morphing. Furthermore, Karungaru~\cite{karungaru2007automatic} 
used an evolutionary algorithm to automatically identify features for morphing faces. 
Our work differs from previously mentioned work in our focus on the direct link between the evolutionary process and image transitions they produce. 

We use well-known random processes for the evolutionary image transition process~\footnote{Images and videos are available at \url{http://cs.adelaide.edu.au/~optlog/research/evol-transitions.php}}. The key idea in this work is to use the evolutionary process {\em{itself}} in an artistic way.

\begin{figure}[t] 
\begin{center}  
\includegraphics[scale=0.3]{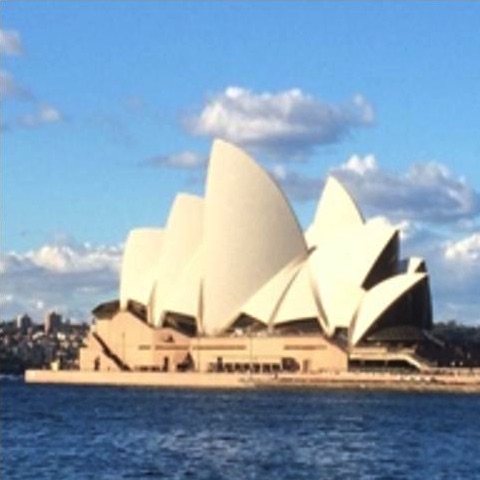} 
\includegraphics[scale=0.3]{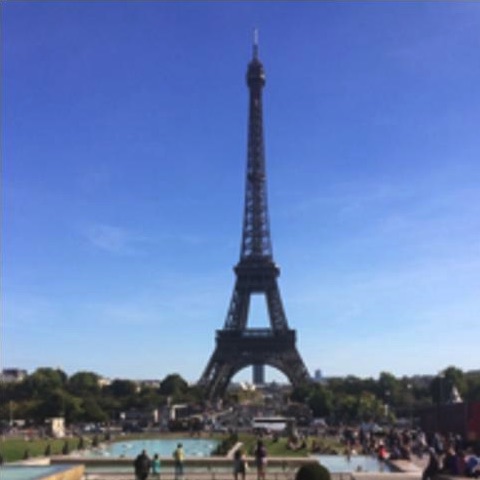} 
\end{center}   
\caption{Starting image $S$ (Sydney Opera House)  and target image $T$ (Eiffel Tower)}
\label{fig:opera}
\end{figure}
The transition process used in this work consists of evolving a given starting image $S$ into a given target image $T$ by random decisions. Considering an error function which assigns to a given current image $X$ the number of pixels where it agrees with $T$ and maximizes this function boils down to the classical OneMax problem for which numerous theoretical results on the runtime behaviour of evolutionary algorithms are available~\cite{DBLP:journals/ec/JansenS10,DBLP:journals/cpc/Witt13,DBLP:journals/tec/Sudholt13}. We use the insights obtained in such studies and show how different processes have an influence on evolutionary image transition.  Furthermore, we are optimistic that the visualization of evolutionary algorithms through image transition may be of independent interest to researchers working on evolutionary computation as it provides a new mechanism of visualizing an evolutionary process. 

Using mutations where in each step exactly one uniformly at random chosen pixel may flip to the target, leads to the Coupon Collector process~\cite{Mitz2005} which significantly slows down when being close to the target. It has been shown in~\cite{DBLP:journals/ec/JansenS10} that an asymmetric mutation operator moves at a constant speed towards the target and thereby avoids the slow-down due to the Coupon Collector's effect. We use a simple (1+1) EA together with this asymmetric mutation operator as our baseline algorithm for evolutionary image transition.
Another important topic related to the theory of evolutionary algorithms are random walks~\cite{Lovasz1996,Dembo2004}. We consider random walks on images where each time the walk visits a pixel its value is set to the value of the given target image. By biasing the random walk towards pixels that are similar to the current pixel we can study the effect of such biases which might be more interesting from an artistic perspective.

After observing these two basic random processes for image transition, we study how they can be combined to give the evolutionary process additional interesting new properties. We study the effect of running random walks for short periods of time as part of a mutation operator in a (1+1) EA. Furthermore, we consider the effect of alternating different mutation operators over time. Our results show that the area of evolutionary image transition based on different well studied random process provides a rich source of artistic possibilities with strong potential 
for further exploration.

The outline of the paper is as follows. In Section~\ref{sec2}, we introduce the evolutionary transition process and examine the behaviour of simple evolutionary algorithms for image transition in Section~\ref{sec:asym}. Section~\ref{sec3} studies how variants of random walks can be used for the image transition process. In Section~\ref{sec4}, we examine the use of random walks as part of mutation operators and study their combinations with pixel-based mutations during the evolutionary process. Finally, we finish with some concluding remarks.

\section{Evolutionary Image Transition}\label{sec2}

\begin{algorithm}[t]

\begin{itemize} 

\item Let $S$ be the starting image and $T$ be the target image.
\item Set X:=S.

\item Evaluate $f(X,T)$.
\item while (not termination condition)
\begin{itemize}
\item Obtain image $Y$ from $X$ by mutation.
\item Evaluate $f(Y, T)$
\item If $f(Y, T) \geq f(X, T)$, set $X:= Y$. 
\end{itemize}
\end{itemize}
\caption{Evolutionary algorithm for image transition}
\label{alg:ea}
\vspace{-0.3cm}
\end{algorithm}

We consider an evolutionary transition process that transforms a given image $X$ of size $m \times n$ into a given target image $T$ of size $m \times n$. Our goal is to study different ways of carrying out this evolutionary transformation based on random processes from an artistic perspective.

We start our process with a starting image $S=(S_{ij})$. Our algorithms evolve $S$ towards $T$ and has at each point in time an image $X$ where $X_{ij} \in \{S_{ij}, T_{ij} \}$. We say that pixel $X_{ij}$ is in state $s$ if $X_{ij} = S_{ij}$, and $X_{ij}$ is in state $t$ if $X_{ij}=T_{ij}$.

 For our process we assume that $S_{ij} \not = T_{ij}$ as pixels with $S_{ij} = T_{ij}$ can not change values and therefore do not have to be considered in the evolutionary process.
To illustrate the effect of the different methods presented in this paper, we consider the Sydney Opera House as the starting image and the Eiffel Tower as the target image (see Figure~\ref{fig:opera}).

The fitness function of an evolutionary algorithm guides its search process and determines on how to move between images. Therefore, the fitness function itself has a strong influence on the artistic behaviour of the evolutionary image transition process. An important property for evolutionary image transition should be that images close to the target image get a higher fitness score. 
We measure the fitness of an image $X$ as the number of pixels where $X$ and $T$ agree. This fitness function is isomorphic to that of the OneMax problem when interpreting the pixels of $S$ as $0$'s and the pixels of $T$ as $1$'s. Formally, we define the fitness of $X$ with respect to $T$ as

\vspace{+0.1cm}
\[
f(X,T) = |\{X_{ij} \in X \mid X_{ij}=T_{ij}\}|.
\]

We consider simple variants of the classical (1+1)~EA in the context of image transition. The algorithm is using mutation only and accepts an offspring if it is at least as good as its parent according to the fitness function. The approach is given in Algorithm~\ref{alg:ea}.
Using this algorithm has the advantage that parents and offspring do not differ too much in terms of pixel which ensures a smooth process for transitioning the starting image into the target.
 Furthermore, we can interpret each step of the random walks flipping a visited pixel to the target outlined in Section~\ref{sec3} as a mutation step which according to the fitness function is always accepted.

\section{Evolutionary Algorithms with Asymmetric Mutation}
\label{sec:asym}

We consider a simple evolutionary algorithm that have been well studied in the area of runtime analysis, namely variants of the classical (1+1)~EA. As already mentioned, our setting for the image transition process is equivalent to the optimization process for the classical benchmark function OneMax.  Our aim is to demonstrate how the progress of these processes are mirrored in the transition of images. The standard variant of the (1+1)~EA  flips each pixel with probability $1/|X|$ where $|X|$ is the total number of pixels in the given image. Using this mutation operator, the algorithm encounters the well-known coupon collector's effect which means that additive progress towards the target image when having $k$ missing target pixels is $\Theta(k/n)$~\cite{DBLP:journals/ai/HeY01}.

\begin{algorithm}[t]
\begin{itemize}
\item Obtain $Y$ from $X$ by flipping each pixel $X_{ij}$ of $X$ independently of the others with probability $c_s/(2|X|_S)$ if $X_{ij}=S_{ij}$, and flip $X_{ij}$ with probability $c_t/(2|X|_T)$ if $X_{ij}=T_{ij}$, where $c_s \geq 1$ and $c_t \geq 1$ are constants.
\end{itemize}
\vspace{-0.3cm}
\caption{Asymmetric mutation}
\label{alg:asym}
\end{algorithm}

In order to avoid the coupon collector's effect, we use the
asymmetric mutation operator introduced and theoretically analyszed in \cite{DBLP:journals/ec/JansenS10}. 
Jansen and Sudholt~\cite{DBLP:journals/ec/JansenS10} have shown that the (1+1)~EA using asymmetric mutation optimizes OneMax in time $\Theta(n)$ which improves upon the usual bound of $\Theta(n \log n)$ when using standard bit mutations. 
In order to apply this asymmetric mutation to our image transition process we make the process of flipping pixels dependent on the number of pixels of $X$ that are in the same state as $X_{ij}$. 
We denote by $|X|_T$ the number of pixels where $X$ and $T$ agree. Similarly, we denote by $|X|_S$ the number of pixels where $X$ and $S$ agree. Each pixel is starting state $s$ is flipped with probability $c_s/(2|X|_S)$ and each pixel in target state $t$ is flipped with probability $c_t/(2|X|_T)$.
The mutation operator is shown in Algorithm~\ref{alg:asym}.

The mutation operator differs from the one given in~\cite{DBLP:journals/ec/JansenS10} by the two constants $c_s$ and $c_t$ which allows the determination of the expected number of new pixels from the starting image  and the target image, respectively. The choice of $c_s$ and $c_t$ determines the expected number of pixel in the starting state and target state to be flipped. To be precise, the expected number of pixel currently in starting state $s$ to be flipped is $c_s/2$ and the number of pixels in target state $t$ to be flipped is $c_t/2$ as long as the current solution $X$ contains at least that many pixel of the corresponding type.
 In \cite{DBLP:journals/ec/JansenS10} the case $c_s = c_t=1$ has been investigated which ensures at each point in time an additive drift of $\Theta(1)$. Using different values for $c_s$ and $c_t$ allows us to change the speed of transition as well as the relation of the number of pixels switching from the starting image to the target and vice versa while still ensuring that there is constant progress towards the target.

All experimental results in this paper are shown for the process of moving from the starting image to the target image given in Figure~\ref{fig:opera} where the images are of size $200 \times 200$. The algorithms have been implemented in MATLAB.
In order to visualize the process, we show the images obtained when the evolutionary process reaches 12.5\%, 37.5\%, 62.5\% and 87.5\% pixels of target image for the first time. We should mention that all processes except the use of the biased random walk are independent of the starting and target image which implies that the use of other starting and target images would show the same effects in terms of the way that target pixels are displayed during the transition process.

For our experiments with \eaasym, we set $c_s=100$ and $c_t=50$ which allows both a decent speed for the image transition process and enough exchanges of pixels for an interesting evolutionary process. We should mention that obtaining the last pixels of the target image may take a long time compared to the other progress steps when using large values of $c_t$. However, for image transition, this only effects steps when the are at most $c_t/2$ source pixels remaining in the image. From a practical perspective, this means that the evolutionary process has almost converged towards the target image and setting the remaining missing target pixels to their target values provides an easy solution. 

\begin{figure}[t] 

\includegraphics[scale=0.2030]{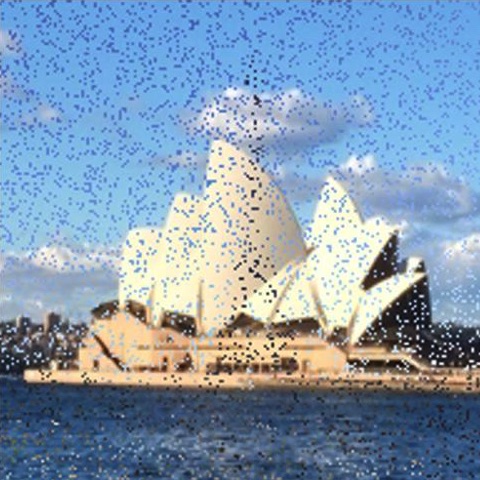}
\includegraphics[scale=0.2030]{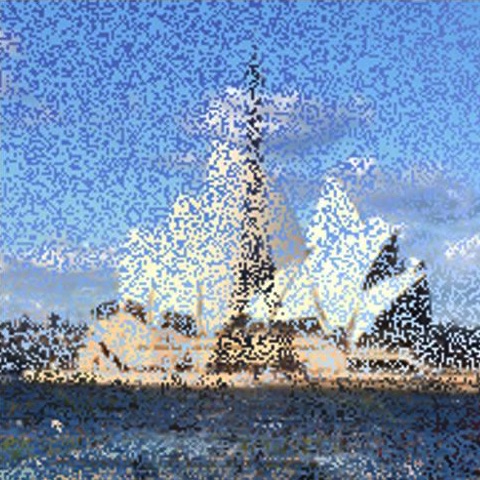}
\includegraphics[scale=0.2030]{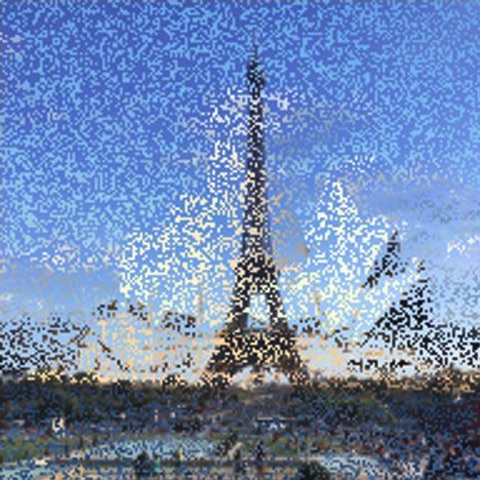}
\includegraphics[scale=0.2030]{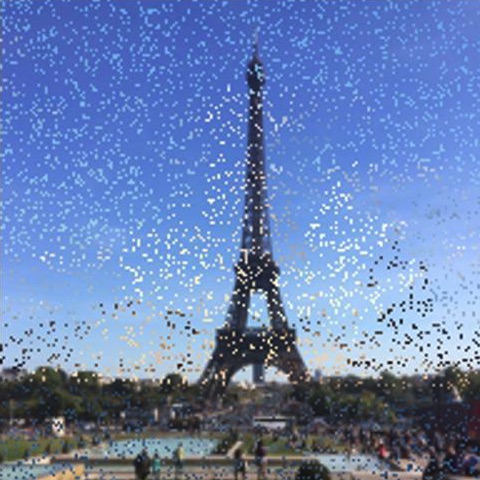}\\

\caption{Image Transition for \eaasym}
\label{fig:2}
\end{figure}

In Figure~\ref{fig:2} we show the experimental results of the asymmetric mutation approach. 
Firstly, we can see the image with lightly stippling dots in randomly chosen areas of the target image $T$. Consequently the area of the white Sydney Opera House disappears and the Eiffel Tower appears. Meanwhile the sky has adopted a dot pattern, where a nuance of dark and light develops steadily. In the last image we barely see the Sydney Opera House, the target image $T$ appearing permanently with the sky becoming darker, whereby the stippling effect shown in the middle two frames decreases.

\section{Random Walks}
\label{sec3}

\begin{algorithm}[t]

\begin{itemize}

\item Let $X_{ij}$ be the starting pixel for the random walk.
\item Set $X_{ij} = T_{ij}$.
\item while (not termination condition)
\begin{itemize}
\item Choose $X_{kl} \in N(X_{ij})$.
\item Set $i=k$, $j=l$, and $X_{ij} = T_{ij}$. 
\end{itemize}
\end{itemize}
\caption{Random walk for image transition}
\label{alg:walk}
\vspace{-0.3cm}
\end{algorithm}

We study random walk algorithms for image transition which moves, at each step, from a current pixel $X_{ij}$ to one pixel in its neighbourhood.

We define the neighbourhood $N(X_{ij})$ of $X_{ij}$ as

\[
N(X_{ij}) = \{ X_{(i-1)j}, X_{(i+1)j}, X_{i(j-1)} X_{i(j+1)} \}
\]

\vspace{-0.1cm}
where we work modulo the dimensions of the image in the case that the values leave the pixel ranges, $i \in \{1, \ldots, m\}$, $j \in \{1, \ldots, n\}$. This implies that from a current pixel, we can move up, down, left, or right. Furthermore, we wrap around when exceeding the boundary of the image.

The classical random walk chooses an element $X_{kl} \in N(X_{ij})$ uniformly at random. We call this the \emph{uniform random walk} in the following. The cover time of the uniform random walk on a $n \times n$ torus is  upper bounded by $4n^2 (\log n)^2/\pi$~\cite{Dembo2004}  which implies that the expected number of steps of the uniform random walk until the target image is obtained (assuming $m=n$) is upper bounded by $4n^2 (\log n)^2/ \pi)$.

We also consider a \emph{biased random walk} where the probability of choosing the element $X_{kl}$ is dependent on the difference in RGB-values for $T_{ij}$ and $T_{kl}$. Weighted random walks have been used in a similar way in the context of image segmentation~\cite{DBLP:journals/pami/Grady06}.
We denote by $T_{ij}^r$, $1 \leq r \leq 3$, the $r$th RGB value of $T_{ij}$ and define

\vspace{-0.4cm}
\[
\gamma (X_{kl}) = \sum_{r=1}^3 |T_{kl}^r - T_{ij}^r|.
\]

\vspace{+0.1cm}

The probability of moving from $X_{ij}$ to an element $X_{kl}  \in N(X_{ij})$ is then given by

\vspace{+0.1cm}
\[
p({X_{kl}}) = \frac{\gamma(X_{kl})}{\sum_{X_{st} \in N(X_{ij})} \gamma(X_{st})}.
\]

\vspace{+0.1cm}
The biased random walk is the only method that is dependent on the target image when carrying out mutation or random walk steps. 
Introducing the bias in terms of pixels that are similar, the bias can take the evolutionary image transition process to take exponentially long as the walk might encounter effects similar to the gambler's ruin process~\cite{Mitz2005}. For our combined approaches described in the next section, we use the random walks as mutation components which ensures that the evolutionary image transition is carried out efficiently.

\begin{figure}[t] 

\includegraphics[scale=0.2030]{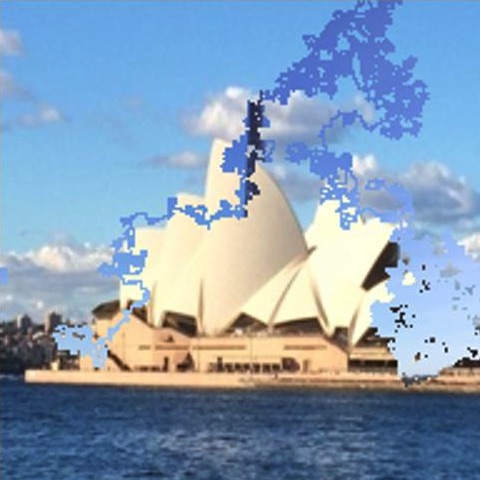} 
\includegraphics[scale=0.2030]{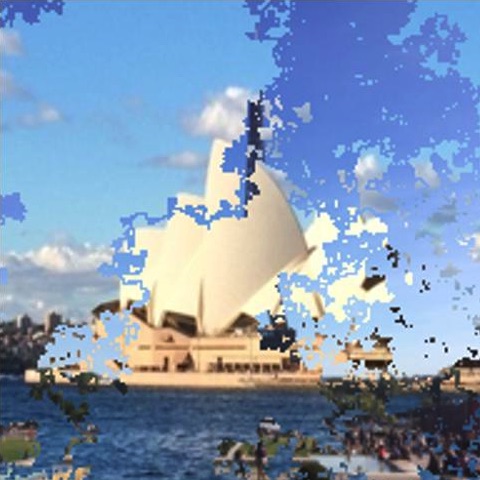}
\includegraphics[scale=0.2030]{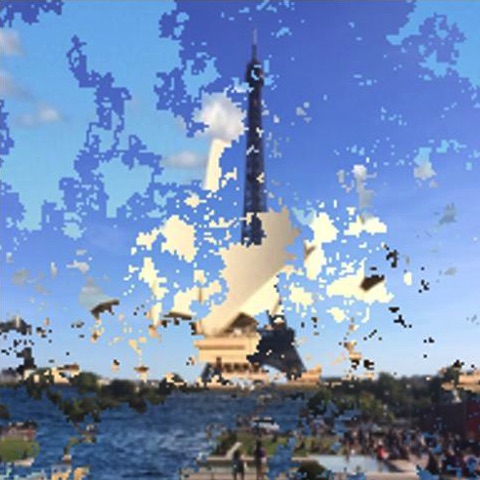} 
\includegraphics[scale=0.2030]{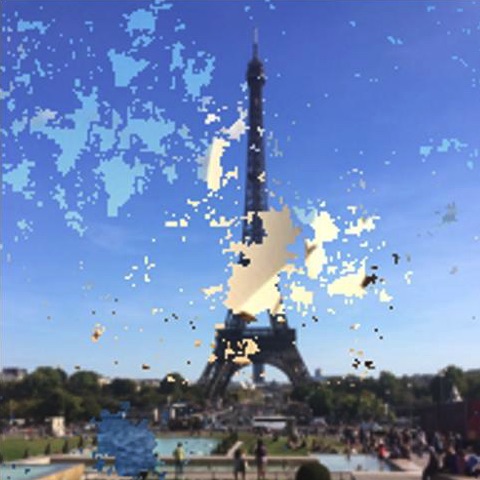}\\ 

\vspace{-0.3cm}
\includegraphics[scale=0.2030]{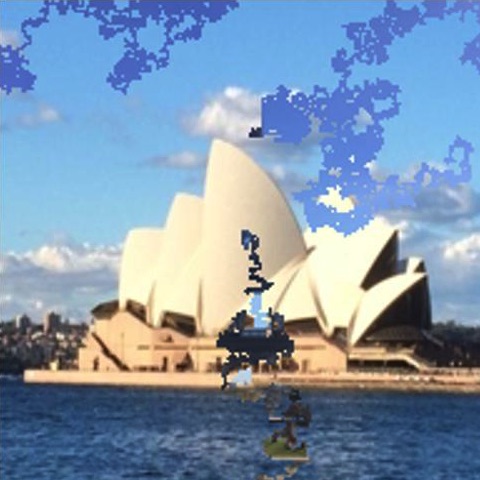}
\includegraphics[scale=0.2030]{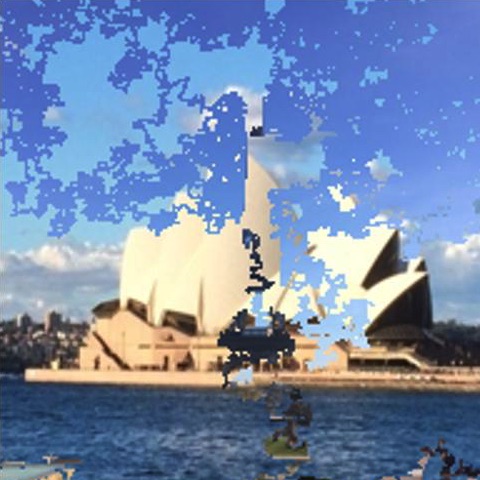}
\includegraphics[scale=0.2030]{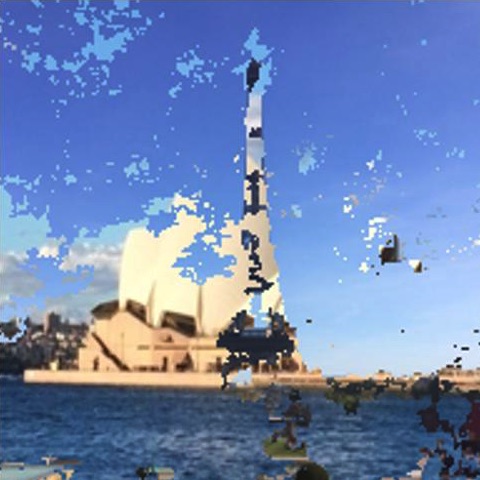}
\includegraphics[scale=0.2030]{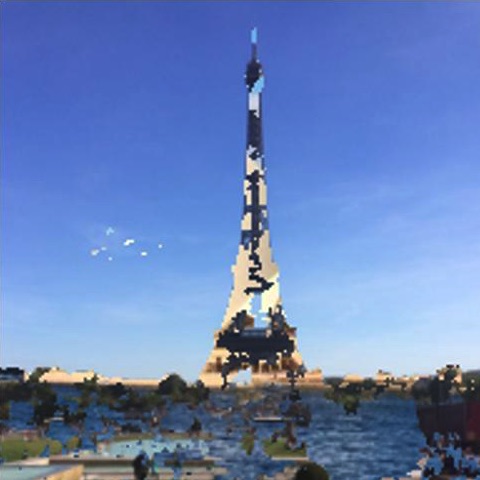}

\caption{Image Transition for Uniform Random Walk (top) and Biased Random Walk (bottom)}
\label{fig:3}
\end{figure}

In Figure~\ref{fig:3} we show the experimental results of the uniform random walk and biased random walk. 
At the beginning, we can observe the image with the characteristic random walk pathway appearing as a patch in the starting image $S$. Through the transition process, the clearly recognisable patches on the target image $T$ emerge. In the advanced stages the darker patches from the sky of the target image dominate. The effect in animation is that the source image is scratched away in a random fashion to reveal an underlying target image.

The four images of the biased random walk are clearly different to the images of the uniform random walk. During the course of the transition, the difference becomes more prominent, especially in the sky, where at 87.5\% pixels of the target image there is nearly an absolute transition to the darker sky. In strong contrast, the lower part of the images stay nearly untouched, so that we see a layer of ocean under the Eiffel Tower. In this image the tower itself is also very incomplete with much of the source picture showing through. These effects arise from biased probabilities in the random walk which makes it difficult for the walk to penetrate areas of high contrast to the current pixel location. 

\section{Combined Approaches}
\label{sec4}

The asymmetric mutation operator and the random walk algorithms have quite different behaviour when applied to image transition. We now study the effect of combining the approaches for evolutionary image transition into order to obtain a more artistic evolutionary process. 

\subsection{Random Walk Mutation}
Firstly, we explore the use of random walks as mutation operators and call this a \emph{random walk mutation}.
The \emph{uniform random walk mutation} selects the position of a pixel $X_{ij}$ uniformly at random and runs the uniform random walk for  $\tmax$ steps. We call the resulting algorithm \rwalk.
Similarly, the \emph{biased random walk mutation} selects the position of a pixel $X_{ij}$ uniformly at random and runs the biased random walk for  $\tmax$ steps. This algorithm is called \bwalk. For our experiments, we set $\tmax=100$.

\begin{figure}[t]

\includegraphics[scale=0.2030]{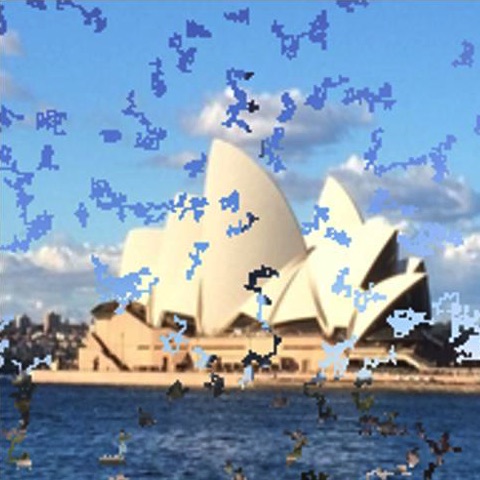} 
\includegraphics[scale=0.2030]{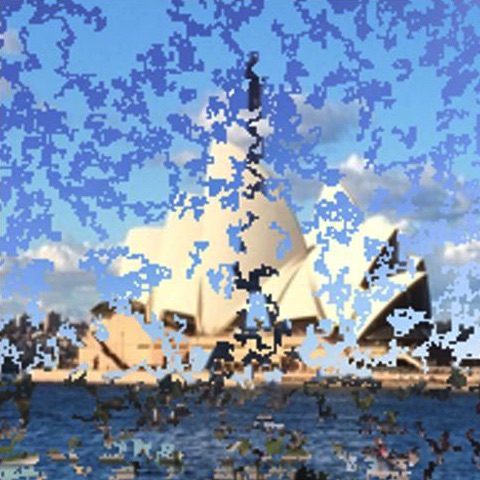}
\includegraphics[scale=0.2030]{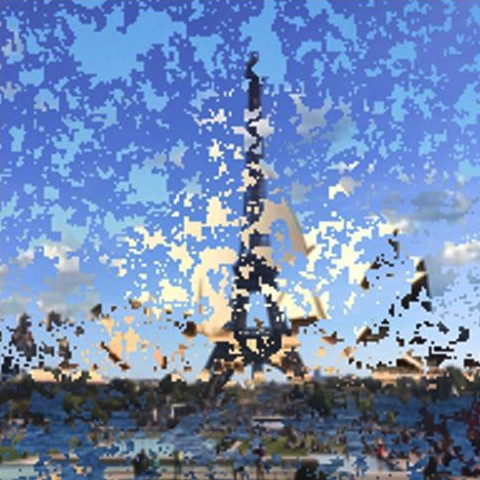} 
\includegraphics[scale=0.2030]{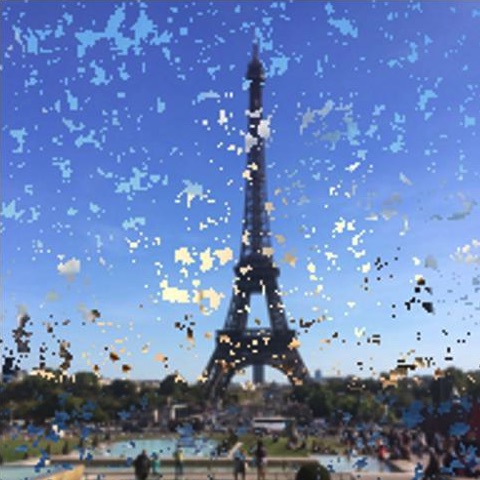}\\

\vspace{-0.3cm}
\includegraphics[scale=0.2030]{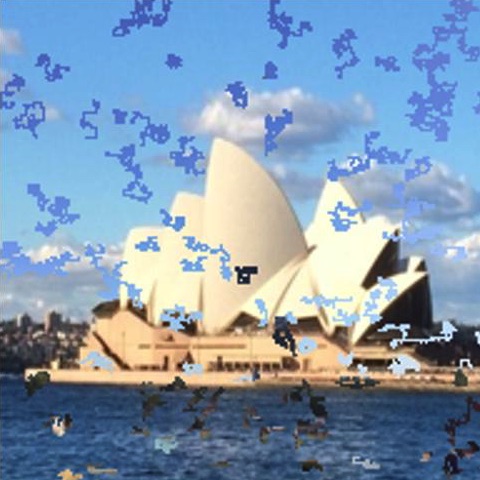}
\includegraphics[scale=0.2030]{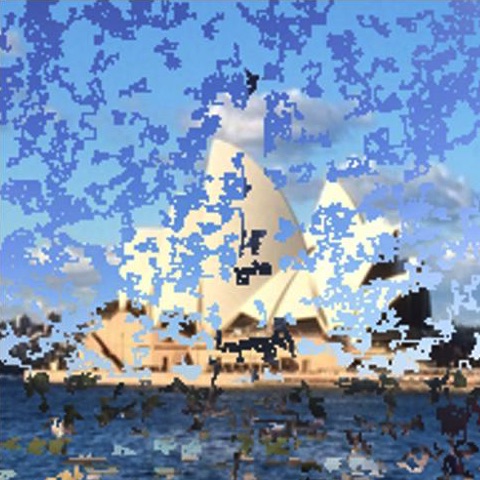}
\includegraphics[scale=0.2030]{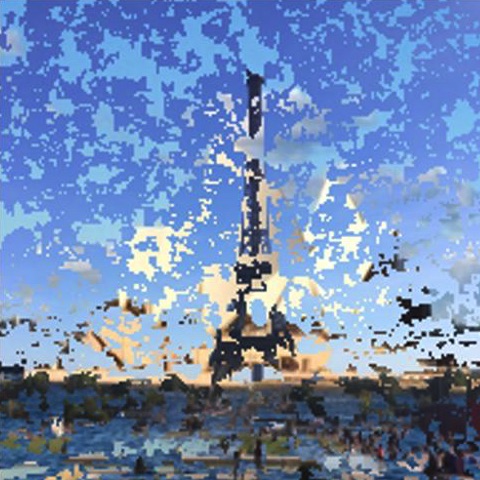}
\includegraphics[scale=0.2030]{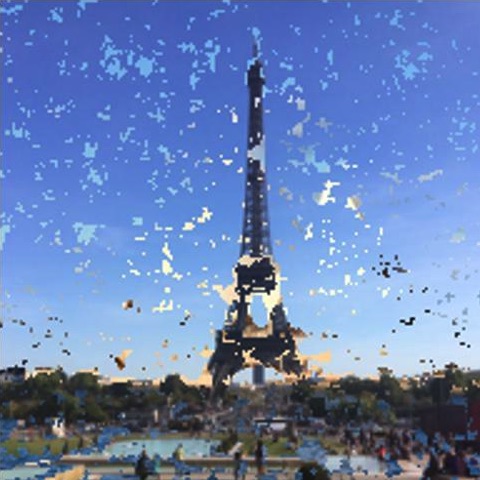}

\caption{Image Transition for \rwalk (top) and \bwalk (bottom)}
\label{fig:4}
\end{figure}

Figure~\ref{fig:4} shows the results of the experiments for \rwalk and \bwalk. The transitions produced were significantly different from the previous ones. In both experiments we can see the target image emerging through a series of small patches at first, then steadily changing through a more chaotic phase where elements of the source and target image appear with roughly equal frequency. On the last image of each experiment we can see most details of the target image. 

The images from \bwalk appear similar to those in \rwalk in the beginning but differences emerge at the final stages of transition where, in \bwalk, elements of the source image still show through in areas of high contrast in the target image, which the biased random walk has difficulty traversing. This mirrors, at a more local scale the effects of bias in the earlier random walk experiments. At a global scale it can be seen that the sky, which is a low contrast area, is slightly more complete in the final frame of \bwalk than the same frame in \rwalk.

\subsection{Combination of asymmetric and random walk mutation}

Furthermore, we explore the combination of the asymmetric mutation operator and random walk mutation.
Here, we run the asymmetric mutation operator as described in Algorithm~\ref{alg:asym} and a random walk mutation every $\tau$ generations. We explore two combinations, namely the combination of the asymmetric mutation operator with the uniform random walk mutation (leading to the algorithm  \arwalk) as well as the combination of the asymmetric mutation operator with the biased random walk mutation (leading to the algorithm \abwalk). We set $\tau=1$ and $\tmax=2000$ which means that the process is alternating between asymmetric mutation and random walk mutation where each random walk mutation carries out $2000$ steps. 

\begin{figure}[t] 

\includegraphics[scale=0.2030]{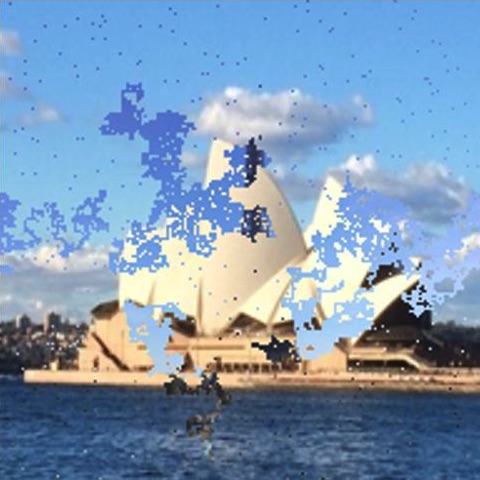}
\includegraphics[scale=0.2030]{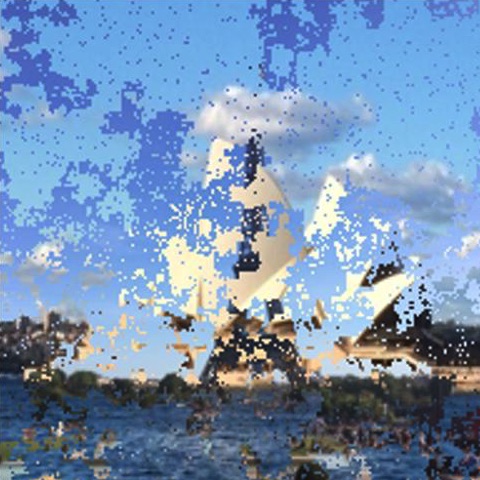} 
\includegraphics[scale=0.2030]{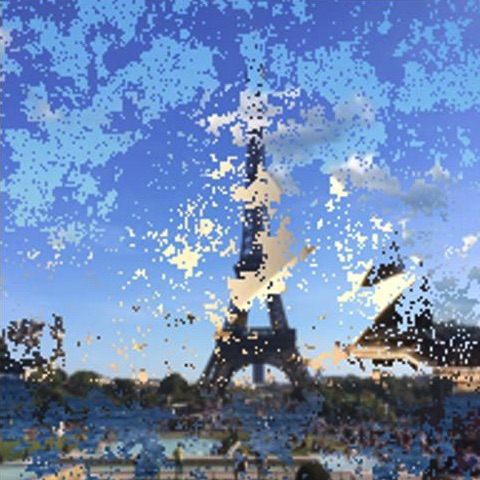}
\includegraphics[scale=0.2030]{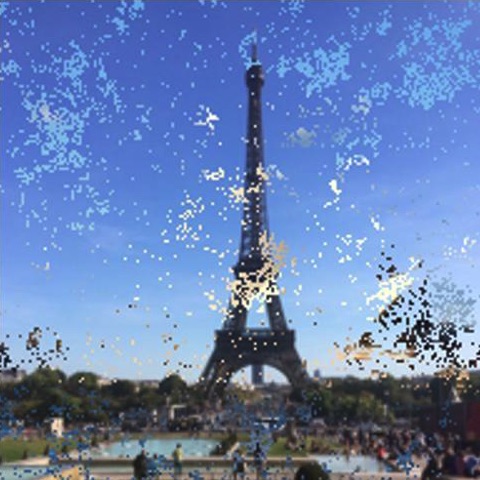} \\

\vspace{-0.3cm}
\includegraphics[scale=0.2030]{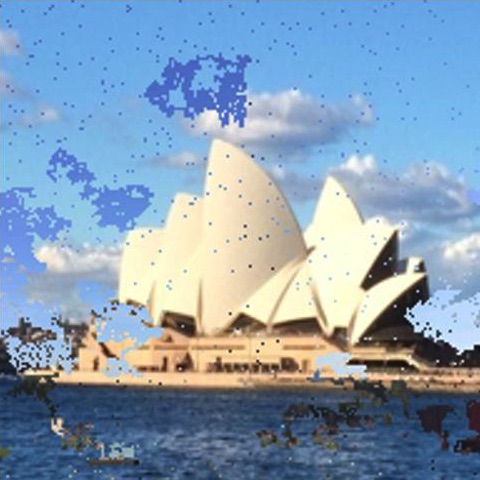}
\includegraphics[scale=0.2030]{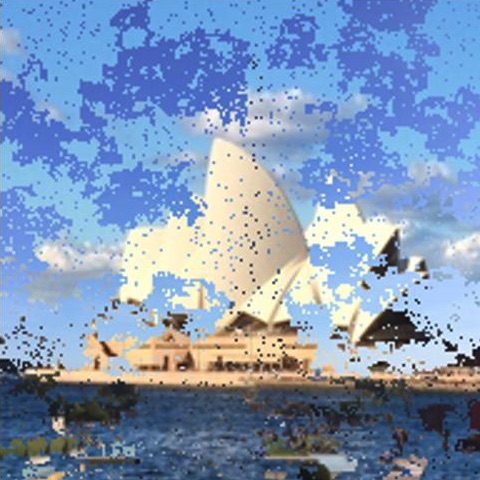}
\includegraphics[scale=0.2030]{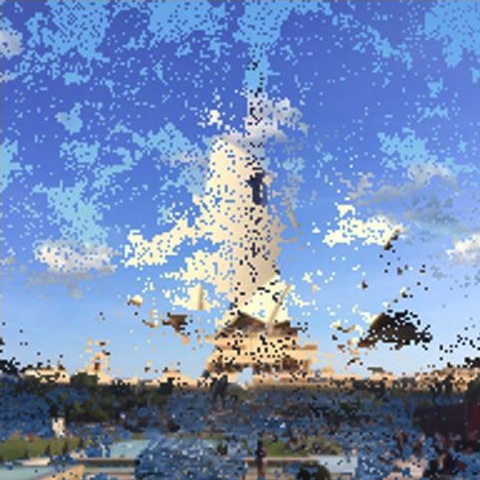}
\includegraphics[scale=0.2030]{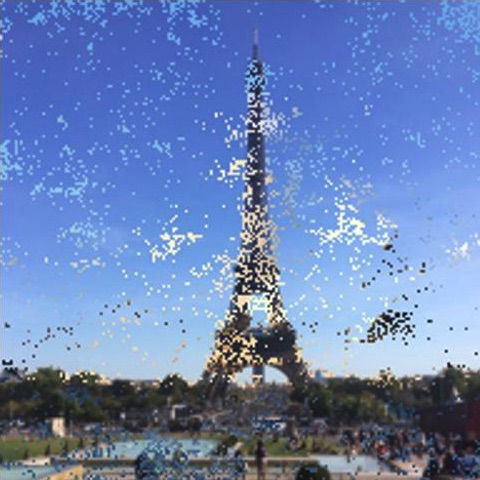}

\caption{Image Transition for \arwalk (top) and \abwalk (bottom)}
\label{fig:5}
\end{figure}

Figure~\ref{fig:5} shows the results of \arwalk  and \abwalk. From a visual perspective both experiments combine the stippled effect of the \eaasym with the patches of the random walk. In \abwalk there is a lower tendency for patches generated by random walks to deviate into areas of high contrast. As the experiment progresses, the pixel transitions caused by \eaasym steps, which have a tendency to degrade contrast barriers, influence this effect. However, even in the final frames there is clearly more sky from the target image in \abwalk than in \arwalk. Moreover, there are more remaining patches of the source image near the edges of the base of the tower, creating interesting effects.

\section{Conclusions and Future Work}

We have investigated how evolutionary algorithms and related random processes can be used for image transition based on some fundamental insights from the areas of runtime analysis and random walks. We have shown how the asymmetric mutation operator introduced originally for the optimization of OneMax can be applied to our problem. Furthermore, we have investigated the use of random walk algorithms for image transition and have shown how they can be used as mutation operators in this context. Investigating the combinations of the different approaches, we have presented various ways of having interesting evolutionary image transition processes.
All our investigations are based on a fitness function that is equivalent to the well-known OneMax problem. For future research it would be interesting to study more complex fitness functions and their impact on the artistic behaviour of evolutionary image transition.

\bibliographystyle{abbrv}

\end{document}